\documentclass[journal]{IEEEtran}
\usepackage{amsmath,amsfonts}
\usepackage{array}

\usepackage{textcomp}
\usepackage{stfloats}
\usepackage{setspace}
\usepackage{url}
\usepackage{verbatim}
\usepackage{graphicx, caption}
\usepackage{cite}
\usepackage{booktabs}

\usepackage[accsupp]{axessibility}  

\usepackage{enumitem}
\usepackage{xspace}
\usepackage{diagbox}
\newcommand{\alg}{NNS-EMD\xspace}

\usepackage{multirow}
\usepackage{algorithm}
\usepackage{subcaption}
\usepackage{algpseudocode}
\usepackage{amssymb}
\newcommand{\Input}[1]{\State \textbf{Input:} #1}
\newcommand{\Output}[1]{\State \textbf{Output:} #1}
\allowdisplaybreaks

\usepackage[pagebackref,breaklinks,colorlinks]{hyperref}
\usepackage{amsthm}
\newtheorem{lemma}{Lemma}
\newtheorem{theorem}{Theorem}

\usepackage[capitalize]{cleveref}
\crefname{section}{Sec.}{Secs.}
\Crefname{section}{Section}{Sections}
\Crefname{table}{Table}{Tables}
\crefname{table}{Tab.}{Tabs.}

\begin{document}

\title{Efficient Approximation of Earth Mover's Distance
Based on Nearest Neighbor Search\thanks{This work was supported in part by ASCENT, one of six centers in JUMP, a Semiconductor Research Corporation (SRC) program sponsored by DARPA, and SUPREME, one of seven centers in JUMP 2.0, a Semiconductor Research Corporation (SRC) program sponsored by DARPA. This work was supported in part by NSF under Grant No. CCF-2212239.}}

\author{\author{Guangyu~Meng\textsuperscript{1}\textsuperscript{\textdagger}, Ruyu~Zhou\textsuperscript{2}\textsuperscript{\textdagger}, Liu~Liu\textsuperscript{1}, Peixian~Liang\textsuperscript{1}, \\
Fang~Liu\textsuperscript{2}, Danny Z.~Chen\textsuperscript{1},
Michael Niemier\textsuperscript{1},
Xiaobo Sharon Hu\textsuperscript{1}

\textsuperscript{1}Department of Computer Science and Engineering, University of Notre Dame, USA \\
\textsuperscript{2}Department of Applied and Computational Mathematics and Statistics, University of Notre Dame, USA\\

\thanks{
\textdagger Equal contribution.}

}

\thanks{This paper was produced by the IEEE Publication Technology Group. They are in Piscataway, NJ.}
}


\markboth{Journal of \LaTeX\ Class Files,~Vol., No.,}%
{Shell \MakeLowercase{\textit{et al.}}: A Sample Article Using IEEEtran.cls for IEEE Journals}


\maketitle

\begin{abstract}
Earth Mover's Distance (EMD) is an important similarity measure between two distributions, commonly used in computer vision and many other application domains. However, its exact calculation is computationally and memory intensive, which hinders its scalability and applicability for large-scale problems. Various approximate EMD algorithms have been proposed to reduce computational costs, but they suffer lower accuracy and may require additional memory usage or manual parameter tuning compared to the exact calculation. In this paper, we present a novel approach, \alg, to approximate EMD using Nearest Neighbor Search (NNS), achieving high accuracy, low time complexity, and high memory efficiency. 
The NNS operation reduces the number of data points processed in each NNS iteration and offers opportunities for parallel processing. We further accelerate \alg via vectorization on GPU, which is especially beneficial for large datasets. We compare \alg with both the exact EMD and state-of-the-art approximate EMD algorithms in image and document classification and image retrieval tasks. We also apply \alg to calculate transport mapping and realize color transfer between images. \alg achieves speed 44$\times$ to 135$\times$ faster than the exact EMD implementation and offers superior accuracy, speedup, and memory efficiency compared to existing approximate EMD methods.

\end{abstract}

\section{Introduction}
\label{sec:intro} 

Earth Mover's Distance (EMD) was first proposed to quantify the similarity between images\cite{rubner1998metric}.
In the theory of Optimal Transport (OT)\cite{villani2009optimal}, EMD is also known as the Kantorovich or Wasserstein-1 distance, and has been widely used in various computer vision (CV) applications and other domains. For instance,  generative modeling of images\cite{wgan,rout2021generative} integrates EMD for improved training quality and stability. The representation of 3D point clouds\cite{urbach2020dpdist,zhao2023earth} and topological shape analysis \cite{chen2021approximation} employed EMD for better capturing nuanced similarities in distributions.
Also, EMD has been adopted for document retrieval in natural language processing (NLP)\cite{kusner2015word,wu-etal-2018-word} due to its flexibility in measuring similarities, even for comparing datasets of different sizes\cite{kusner2015word,ling2007efficient}.

Although EMD offers highly effective similarity measurements for images and other types of high-dimensional data, computing the exact EMD relies on linear programming (LP)-based algorithms, such as the Hungarian method \cite{kuhn1955hungarian}, the auction algorithm \cite{alaya2019screening}, and network simplex \cite{waissi1994network}. These LP methods are numerically robust but still face significant challenges: \textbf{(i)} computationally intensive LP-based implementations \cite{kusner2015word}, which can incur $O(n^3 \log n)$ time complexity \cite{shirdhonkar2008approximate}; \textbf{(ii)} the need to store the results of exhaustive comparisons between data points, which can incur high memory footprints \cite{liu2020morphing}, restricting the exact EMD from being employed in applications that require fast response in large-scale data settings (e.g., visual tracking \cite{yao2018visual}, online image retrieval \cite{xie2017dynamic}). 
    
    

\begin{figure}[bt]
    \centering
    \hspace*{-0.15in}
    \includegraphics [scale = 0.12]{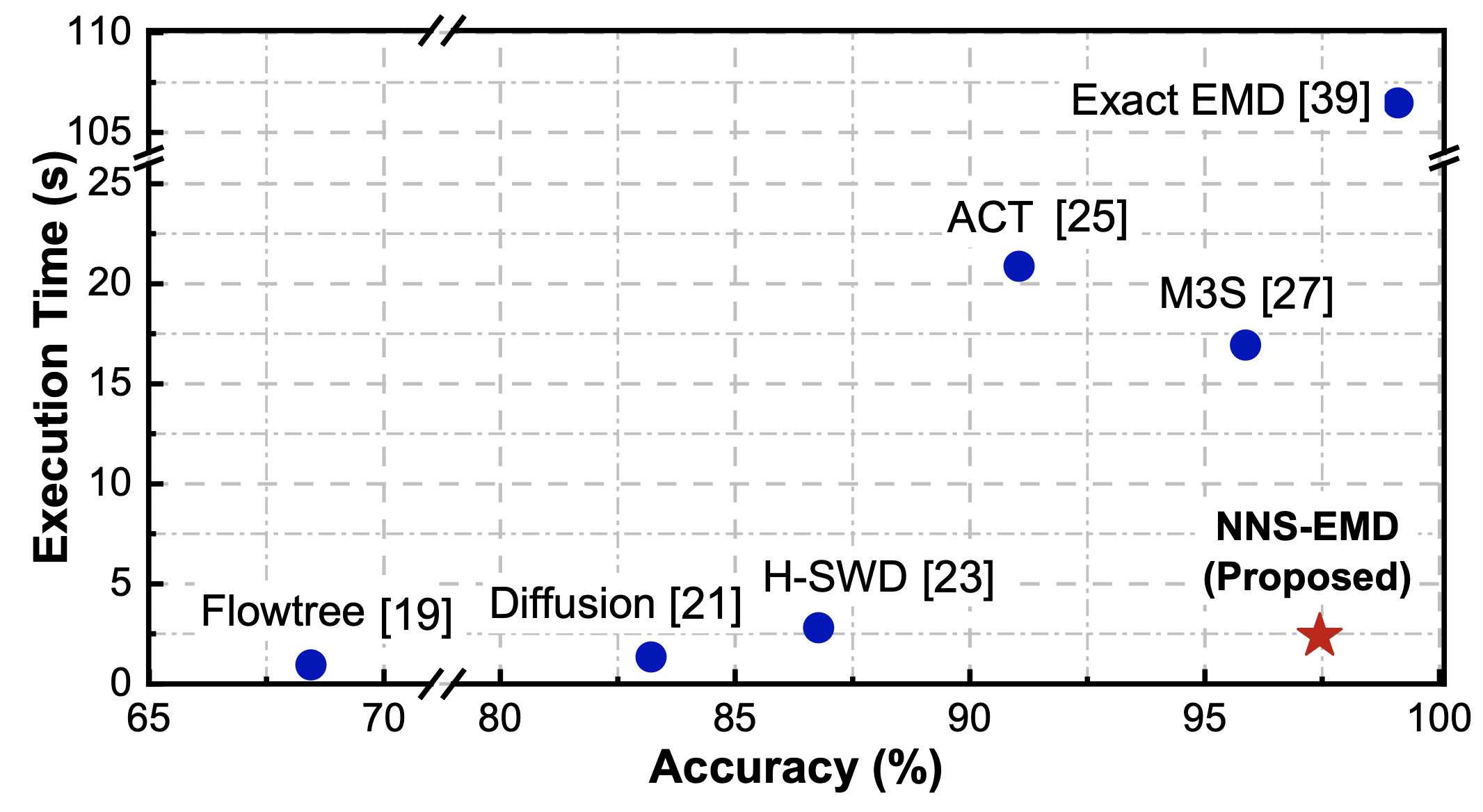}
    
    \caption{Execution time ($y$-axis) versus classification accuracy ($x$-axis) among exact EMD and SOTA approximate EMD algorithms for the MNIST dataset \cite{lecun1998gradient}.}
    \label{fig:motivation}
    
\end{figure}

A variety of approximate algorithms for EMD have been proposed to address the high computational overhead of computing the exact EMD~\cite{tree,le2019tree,tong2021diffusion,le2022sobolev,nguyen2023hierarchical,nguyen2021point,atasu2019linear,kusner2015word}. Some algorithms employ pre-defined data structures (e.g., trees or graphs) to store the entire datasets offline\cite{tree,le2019tree,tong2021diffusion,le2022sobolev}. 
Leveraging auxiliary data structures, these algorithms can achieve linear time complexity at the expense of additional memory space usage and lower accuracy. Other types of EMD approximations use entropic regularization \cite{cuturi2013sinkhorn, alaya2019screening, chen2022computing}, dimensional reduction \cite{bonneel2015sliced,nguyen2021point, nguyen2023hierarchical}, or constraints relaxation for optimal transport \cite{atasu2019linear,kusner2015word} to reduce the computation overhead, but also suffer various degrees of accuracy loss. Additionally, the performance of these approximation methods depends on parameter tuning.

In this paper, we propose \alg, a new approximate EMD algorithm that leverages nearest neighbor search (NNS). \alg achieves both high accuracy and relatively low execution time compared to the exact EMD and the existing state-of-the-art (SOTA) approximate EMD algorithms, as shown in Fig.~\ref{fig:motivation}. 
Instead of evaluating all possible point-to-point distances as in the exact EMD calculation, our proposed \alg focuses on identifying the nearest neighbors.
This approach significantly reduces the number of data points being compared and usually does not involve a linear number of iterations in experiments. Note that neighbor distances not only have been theoretically proven as being close approximations of the global distance between two images \cite{borgefors1984distance}, but are also are extensively employed in the 3D point cloud domain \cite{wu2021density,nguyen2021point,lin2023hyperbolic}. The main contributions of our work are as follows:

\begin{itemize}
    \item We propose an efficient EMD approximation, \alg, based on nearest neighbor search, to achieve both high accuracy and low execution time. 

    \item We provide a theoretical analysis of the time complexity and error bound of the proposed \alg.

    \item We introduce a highly parallel implementation using GPU vectorization to further accelerate \alg.
\end{itemize}

We evaluate \alg for classification and retrieval tasks in the CV and NLP domains, and compare it with the exact EMD, other SOTA approximate EMD algorithms, and deep learning (DL) methods with respect to accuracy and execution time. We also compare the memory usage of \alg with other SOTA approximate EMD algorithms. As NNS-EMD can generate transport mappings to perform color transfer across images (a task that is widely used in CV), we also assess the performance of \alg in color transfer using both qualitative and quantitative measures.


Our experimental results show that \alg can offer speedups of 44$\times$ to 135$\times$ versus an exact EMD implementation on the image classification task. Compared to SOTA EMD approximations, \alg obtains up to 5\% higher accuracy with similar latencies for classification and retrieval tasks in the CV and NLP domains. Our code is available at \url{https://github.com/gm3g11/NNS-EMD}.

\section{Background}

This section reviews the definition and mathematical formulation of EMD and then discusses SOTA algorithms for approximating EMD. 

\subsection{EMD and Optimal Transport}\label{sec:emd_bg}

EMD is a mathematical metric for measuring the similarity between two probability distributions. It can be intuitively interpreted as the minimum total cost for transforming one probability distribution to another. We focus on the case where the two probability distributions are discrete.  Let $S$ and $C$ denote two discrete probability distributions on a finite metric space $\mathcal{M} = (X, d_X)$ ($S$ stands for ``suppliers'' and $C$ stands for ``consumers''). $S$ and $C$ can be formulated as two histograms, where $s_i$ is the weight of the $i$-th bin of $S$ and $c_j$ is the weight of the $j$-th bin of $C$. Suppose  $S$ and $C$ have $m$ and $n$ bins with positive weights and both distributions are normalized (i.e., $\sum_i s_i = \sum_j c_j = 1$). The EMD between $S$ and $C$ is defined as:
\begin{align}
    \mbox{EMD}(S, C) &= \min_{F_{i,j}\geq 0}\textstyle\sum_{i,j}F_{i,j}D_{i,j}\label{objective}\\
    \text{subject to: }&\textstyle\sum_i F_{i,j} = c_j \label{inflow},\\
    &\textstyle\sum_j F_{i,j} = s_i,  \label{outflow}
\end{align}
where  $\mathbf{D}$  is a known $m \times n$ non-negative symmetric distance matrix, and $\mathbf{F}$ is a non-negative $m \times n$ flow matrix to be optimized (each entry $F_{i,j}$ in $\mathbf{F}$ represents the amount of weight that flows from the $i$-th bin of $S$ to the $j$-th bin of $C$ in the transformation). The solution for $\mathbf{F}$ is optimized for the minimum total cost of transforming $S$ to $C$.

In the context of measuring similarity between images, image pixels can be formulated as bins in the histograms on which EMD is calculated. Each bin in a histogram is associated with a weight (e.g., $s_i$ or $c_j$) and a pixel coordinate based on which $\mathbf{D}$ can be calculated. $L_1$ and $L_2$ distances are popular choices for $D_{i,j}$ in the image domain. 

There are two common types of histogram formulation for images: the first is spatial location-based histogram \cite{tree,atasu2019linear,cuturi2013sinkhorn, chen2022computing}, and the second is color intensity-based histogram \cite{rubner2000earth,chen2022computing}. In the former, every pixel represents one histogram bin, its weight is the normalized pixel value, and its coordinate is based on the pixel position. In the latter formulation, the number of bins is given by the number of distinct pixel values, and the weight of a bin is the normalized frequency of its pixel value and its coordinate is represented by the RGB values of the pixel.

\subsection{Related Work}
In the realm of EMD approximations, various innovative methods have been proposed. The Flowtree method \cite{tree} embeds a dataset into a tree structure to facilitate the matching of queries in a hierarchical manner. The Diffusion method \cite{tong2021diffusion} diffuses distributions across multi-scale graphs, comparing diffused histograms to approximate EMD. The hierarchical sliced Wasserstein distance (H-SWD) method \cite{nguyen2023hierarchical} implements a hierarchical scheme for projecting datasets into one-dimensional distributions, enhancing efficiency for high-dimensional data. The Approximate Iterative Constrained Transfers (ACT) method \cite{atasu2019linear} improves R-WMD \cite{kusner2015word} by imposing relaxed constraint \eqref{inflow} through iterations. The Multi-scale Sparse Sinkhorn (M3S) method \cite{chen2022computing} partitions a dataset hierarchically into subsets, integrating a multi-scale approach with the Sinkhorn algorithm for EMD approximation. 
Additionally, \cite{huguet2023geodesic} 
accelerates the Sinkhorn algorithm using sparse graph Laplacian for EMD computation with geodesic ground distance. \cite{forrow2019statistical} first introduces low-rank constraints on the feasible couplings in OT problems, with an algorithm tailored for the squared Euclidean ground distance. Building on this, \cite{scetbon2021low} proposes a generic linear-time solver for OT problems with arbitrary costs under low-rank constraints.
Beyond the methods above, another line of work, originating from the geometric transportation problem, has introduced near-linear time approximation schemes with strong theoretical guarantees \cite{Kyle2022, Agarwal2017FasterAlgorithms, sharathkumar2012algorithms} that are also applicable for approximating EMD.

\section{Methodology}

In this section, we present our NNS-EMD algorithm and discuss key insights and challenges for each stage. We also perform theoretical analysis for the error bounds of the \alg solution and its time complexity. Lastly, we conduct an in-depth examination of \alg and further accelerate it by leveraging GPU parallelism.

\subsection{\alg Algorithm} \label{sec:nns-emd}

\begin{figure*}[bt]
    \centering
    \includegraphics [scale = 0.21] {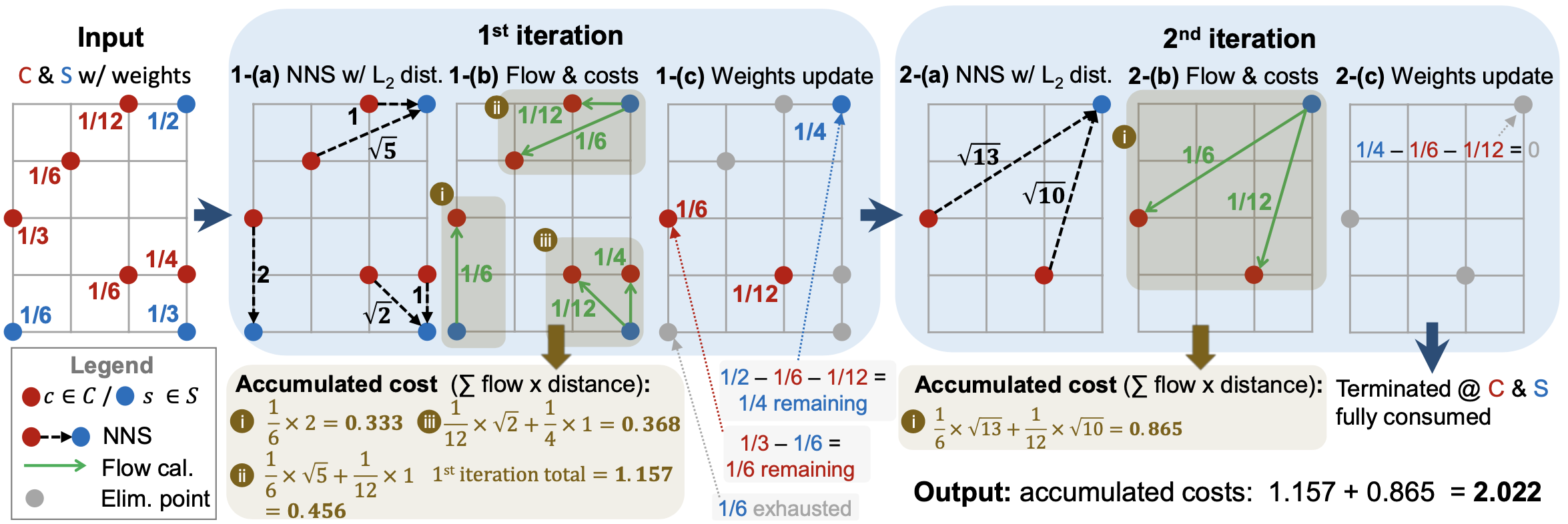}
    \caption{ A toy example illustrating the execution flow of \alg. The inputs are $C$ and $S$, two sets of data points with weights and position coordinates (e.g., pixels in two images). ``w/ '' denotes ``with''. In steps 1-(a) and 2-(a), \alg identifies the nearest neighbor (NN) in $S$ for each point in $C$ to form groups based on the $L_2$ distance. Steps 1-(b) and 2-(b) compute flows between the NN pairs in $S$ and $C$ and calculate the cost for each group. Steps 1-(c) and 2-(c) update the weight for each data point and eliminate the points with zero weight.}
    \label{fig:nns-emd_flow}
\end{figure*}

The inputs to \alg are two sets $S$ and $C$ of data points (e.g., pixels in images) with coordinates and normalized weights, and the output is an approximate EMD (i.e., the total cost for transforming $S$ to $C$). The iterative algorithm consists of 3 key steps: \textbf{NNS operation}, \textbf{flow \& cost calculation}, and \textbf{weights update}. A schematic diagram of the algorithm is given in Fig.~\ref{fig:nns-emd_flow} (for demonstration purposes, it only displays the first two iterations). The algorithm keeps running until convergence, defined as: the total weight of all the points in $S$ and $C$ becomes zero simultaneously.
The location-based histogram described in Sec.~\ref{sec:emd_bg} is employed here for illustration, and the NNS is based on the $L_2$ distance between two position coordinates from $S$ and $C$. 

Below, we explain each of its three key steps and their motivations, with full details of \alg shown in Alg.~\ref{alg:nns}.

\textbf{NNS Operation.} The NNS operation identifies the supplier in $S$ that is the closest for every consumer point from $C$ according to the $L_2$ distance (Fig.~\ref{fig:nns-emd_flow} 1-(a) and 2-(a)). There are two advantages of using NNS: \textbf{(i)} Compared to known data structure-based methods (e.g., tree and graph structures), NNS provides a more direct spatial proximity between two points and can better preserve local structures as no embedding of the original data points into a data structure is needed. \textbf{(ii)} Compared to the LP-based exact EMD computation, NNS-EMD can reduce time complexity by calculating transportation costs only between the nearest neighbor pairs instead of evaluating all possible pairs of points. Furthermore, we analyze the NNS operation and develop a near-linear time GPU implementation in Sec.~\ref{sec:improvement}. This implementation avoids the quadratic time complexity associated with repeated distance evaluations between every pair of data points in $S$ and $C$, thereby enhancing computational efficiency. Note that the choice of distance metric is pivotal for the NNS operation. We examine both the $L_1$ and $L_2$ distances in the experiments (Sec.~\ref{sec:distance}), and $L_2$ distance-based implementation achieves better accuracy and robustness. Thus, we adopt the $L_2$ distance in the rest of the paper.

\textbf{Flow \& Cost Calculation.} In each iteration, after pairing every point in $C$ with its NN in $S$, the algorithm calculates the flow and transportation cost (Fig.~\ref{fig:nns-emd_flow} 1-(b) and 2-(b)). If the $i$-th supplier from $S$ is matched with only the $j$-th consumer in $C$ as its NN, flow $F_{i,j}$ from the former to the latter is defined as $\min\{s_i, c_j\}$. Note that $s_i$ and $c_j$ are the current weight stored in $i$-th supplier and the $j$-th consumer respectively, as they both need to be updated in the next step for every iteration. 

Next, we will discuss how a supplier assigns its weight when is connected by multiple consumers. When multiple consumers share the same NN (e.g., the $i$-th supplier from $S$ is identified as the NN by both the $j_1, j_2$-th consumers from $C$), and if $s_i \geq c_{j_1} + c_{j_2}$
(i.e., the weight $s_i$ stored in the $i$-th supplier can meet the demands of all these consumers), we allocate $s_i$ to each consumer, possibly with some remaining weight after this allocation. When $s_i$ is smaller than the sum of the weights stored in these consumers (the supplier cannot meet all the demands), we need to design a protocol to ``optimally'' allocate the ``limited'' $s_i$. 

One intuitive scheme, named Random Protocol (RP), is to randomly assign $s_i$ to these consumers until $s_i$ is depleted. Alternatively, we may use a Greedy Protocol (GP) that assigns priorities to the matched consumers based on the distance matrix $\mathbf{D}$, with a closer consumer having a higher priority to receive from supplier $i$. For example, if $D_{i,j_1}\leq D_{i,j_2}$, we first calculate $F_{i, j_1} = \min \{s_i, c_{j_1}\}$ and update $s_i$ before calculating $F_{i, j_2}$. The GP could be more time-consuming than the RP as it involves an additional step of ranking the distances of the consumers that are paired with the $i$-th supplier. Yet, the GP is more conceptually consistent with the exact EMD and is expected to yield a more accurate EMD approximation. We implement both the protocols in our experiments, and the results suggest that the GP gives higher accuracy at the expense of increased execution time (see discussions below).

After the flow values $F_{i,j}$ are determined, the algorithm computes the transportation cost for each flow, i.e., $F_{i,j}\times D_{i,j}$, where $D_{i,j}$ is an entry in the distance matrix $\mathbf{D}$. The total transportation cost in an iteration is $\sum_{i,j}(F_{i,j}\times D_{i,j})$.

\vspace{2pt}
\textbf{Weights Update.} In this step, the weights of all the points in both $S$ and $C$ that are involved in the NNS and the flow calculation steps are updated (Fig.~\ref{fig:nns-emd_flow} 1-(c) and 2-(c)). Specifically, $s_i \leftarrow \max\{0,s_i-\sum_j F_{i,j}\}$ and $c_j \leftarrow \max\{0, c_j- F_{i,j}\}$, provided that one supplier may have multiple consumers whereas each consumer has only one supplier. When a point's weight is fully exhausted or consumed, it is eliminated, and only the points with positive weights will participate in the next iteration. Given that both weights in $S$ and $C$ are normalized to 1 and iteratively reduced by the same flow amount, total weights of all points in $S$ or $C$ will reach zero simultaneously. The weights update operation not only ensures the convergence of the NNS-EMD algorithm until all the points are eliminated, but also contributes to efficient computation with a continuously reduced set of data points entering subsequent iterations of the algorithm.

\begin{algorithm}[t]
\caption{NNS-EMD $(S, C, \mathbf{D})$ with Greedy Protocol}
\label{alg:nns}
\begin{algorithmic}[1]
    \Input{Histograms $S$ and $C$, and distance matrix $\mathbf{D}$}
    \State $R=0$ \Comment{initialize accumulated cost $R$}
    \State $\mathbf{F}^* \!\!= [0]_{m\times n}$ \Comment{initialize the flow matrix $\mathbf{F}^*$}
    \State $t=1$ \Comment{initialize the index of NNS iteration $t$}
    \While{$\exists c_j > 0$ or $\exists s_i > 0$}\Comment{while $\exists$ non-zero weight}
        \vspace{2pt}
        \State $\mathbf{m}^{(t)}\! = \{i \ | \ s_i > 0 \}$\Comment{update indices for $S$}
        \vspace{2pt}
        \State $\mathbf{n }^{(t)}\!= \{j \ | \ c_j > 0 \}$\Comment{update indices for $C$}
            \For{$j \in \mathbf{n}^{(t)}$} \Comment{iterate for every active $j$ in $C$}
            \vspace{2pt}
            \State $\mbox{NNS}(j)\!\! =\! \mbox{argmin}_{i\in \mathbf{m}^{(t)}}\!\{D_{i,j}\}$ \label{step:NNS}
            \Comment{NNS in $S$}
            \EndFor
            \For{$i \in \mathbf{m}^{(t)}$}\Comment{iterate for every active $i$ in $S$}
                \vspace{2pt}
                \State $\mathbf{a}_i^{(t)} \!\!= \!\{ j \ | \ c_j > 0, \mbox{NNS}(j) \!=\! i \}$
                \vspace{2pt}
                \State $\widetilde{\mathbf{a}}_i^{(t)}\!\!=\mbox{argsort}(D_{i, \mathbf{a}_i^{(t)}}\!)$ \Comment{sort indices by dist. val.}\label{step:argsort}
                \State $l = 1$ \Comment{initialize $l$}
                \While{$s_i > 0$ and $l < |\widetilde{\mathbf{a}}^{(t)}_i|$}
                \State $f = \min (s_i, c_{\widetilde{\mathbf{a}}^{(t)}_i[l]})$ \Comment{max weight to flow}\label{step:f}
                \State $s_i = s_i - f$ \Comment{update weight of $s_i$}
                \vspace{2pt}
                \State $c_{\widetilde{\mathbf{a}}^{(t)}_i[l]} \!\!= \!c_{\widetilde{\mathbf{a}}^{(t)}_i[l]} \!\!- \!f$ \Comment{update weight of $c_{\widetilde{\mathbf{a}}^{(t)}_i[l]}$}
                \vspace{5pt}
                \State $R = R + f\cdot D_{i, \widetilde{\mathbf{a}}^{(t)}_i[l]}$ \Comment{accumulate cost}
                \vspace{2pt}
                \State $F^*_{i,\widetilde{\mathbf{a}}^{(t)}_i[l]}= F^*_{i,\widetilde{\mathbf{a}}^{(t)}_i[l]} + f$ \Comment{update $\mathbf{F}^*$}
                \vspace{2pt}
                \State $l = l + 1$ \Comment{increase $l$}
                \EndWhile
            \EndFor
        \State $t = t + 1$ \Comment{increase $t$}
    \EndWhile
    \Output{$R$, $\mathbf{F}^*$}
\end{algorithmic}
\end{algorithm}

\subsection{Error Analysis}

We provide a theoretical analysis of the error bound between the \alg solution (output $R$) via Alg.~\ref{alg:nns} and the exact EMD in Lemma~\ref{lemma}, Theorems~\ref{theorem:bound} and \ref{theorem:exact}. The detailed proofs of the results are given in the Supplementary Materials.  

The notation used in the theoretical analysis are as follows. In the $t$-th iteration of Alg.~\ref{alg:nns}, $\mathbf{m}^{(t)}$ and $\mathbf{n}^{(t)}$ contain the indices of the bins with positive weights in $S$ and $C$, respectively, $\mathbf{a}_i^{(t)}$ denotes the indices of the bins in $C$ that identify the $i$-th supplier from $S$ as NN, and $\smash{\widetilde{\mathbf{a}}_i^{(t)}\!\!=\!\mbox{argsort}(D_{i, \mathbf{a}^{(t)}_i})}$ is the sorted version of $\mathbf{a}_i^{(t)}$ according to the distance to the $i$-th supplier. 

Lemma~\ref{lemma}  demonstrates that $\mathbf{F}^*$ is a feasible solution to \eqref{objective}, satisfying constraints \eqref{inflow} and \eqref{outflow} in Sec.~\ref{sec:emd_bg}.

\begin{lemma}[\textbf{$\mathbf{F}^*$ is a feasible solution}]\label{lemma}
Assume for an $i \!\in\!\{1,2, \ldots, m\}$, the $i$-th supplier from $S$ has at least one matched consumer for $\Delta$ iterations for an integer $\Delta\!>\!0$. Without loss of generality, assume $i \!\in\! \mathbf{m}^{(t_1)}\cap \mathbf{m}^{(t_2)}\cap \cdots \cap \mathbf{m}^{(t_\Delta)}$ with $t_1\!<\!t_2\!<\!\cdots\!<\!t_\Delta$, and for $u\!\in \!\{t_1,t_2, \ldots, t_\Delta\}, |\mathbf{a}_i^{(u)}|\geq1$ (in other words, the $i$-th supplier is first identified in the $t_1$-th iteration as an NN to some point in $C$ and its weight is not exhausted until the $t_\Delta$-th iteration). Then, 
\begin{equation}\textstyle\sum_{u=t_1}^{t_\Delta}\sum_{l=1}^{|\mathbf{a}^{(u)}_i|}F^*_{i, \widetilde{\mathbf{a}}^{(u)}_i[l]} = s_i,
\end{equation}
implying that $\mathbf{F}^*$ is a feasible solution to \eqref{objective} satisfying constraints \eqref{inflow} and \eqref{outflow} in Sec.~\ref{sec:emd_bg}.
\end{lemma}

The proof of Lemma~\ref{lemma} follows by keeping track of the flow quantities $f$ in step~\ref{step:f} of Alg.~\ref{alg:nns}.
And based on Lemma~\ref{lemma}, we can derive an upper-bound for the total transport cost difference between $\mathbf{F}^*$ (from Alg.~\ref{alg:nns}) and any feasible flows $\mathbf{F}$ for one individual supplier, and the difference between the EMD solution $R$ from Alg.~\ref{alg:nns} and the exact EMD in Theorem~\ref{theorem:bound}. Furthermore, building on an additional assumption, we can further establish the exactness of $R$ as an approximation to the exact EMD as stated in Theorem~\ref{theorem:exact}.

\begin{theorem}[\textbf{Error bound of EMD approximation by NNS-EMD}] \label{theorem:bound}
Let $\mathbf{F}$ be any feasible flows, and $\mathbf{F}^{opt}$ be the optimal flow solution to \eqref{objective} satisfying constraints \eqref{inflow} and \eqref{outflow}. 
For the $i$-th supplier from $S$ defined in Lemma \ref{lemma}, let $J_i$ be the indices of all consumers that have flow with $s_i$ and $j^*_i$ be the index of the furthest consumer from $C$ that receives a positive flow from $s_i$. Then, 
\begin{equation}\label{eqn:bound}
    \textstyle\sum_{j}\!F^*_{i,j}D_{i,j} - \sum_j\! F_{i,j}D_{i,j}\leq \: D_{i, j^*_i}\!\left(\sum_{j\in J_i}(c_j - F_{i,j})\right) ,
\end{equation}
and the difference between $R$ (the approximate EMD from Alg.~\ref{alg:nns}) and the optimal EMD is:
\small
\begin{align}\label{eq:R-EMD}
\textstyle 0\leq \sum_{i,j}\!F^*_{i,j}D_{i,j} \!-\! \sum_{i,j}\! F^{opt}_{i,j}D_{i,j} \!\leq\! \sum_i\! D_{i, j^*_i}\!\!\left(\sum_{j\in J_i}(c_j \!-\! F^{opt}_{i,j})\right).\notag
\end{align}
\end{theorem}

Theorem~\ref{theorem:bound} suggests that for the $i$-th supplier, the transport cost difference between $\mathbf{F}^*$ and $\mathbf{F}$ is not greater than the distance to the furthest consumer to whom $s_i$ supplies a positive flow, multiplied by the cumulative discrepancies between the original consumer's demand $c_j$ and the actual flow $F_{i,j}$ from $\mathbf{F}$ across all consumers $j \in J_i$ (see Eq.~\eqref{eqn:bound}). Furthermore, the difference between the approximate EMD from \alg and the optimal EMD can be bounded by the sum of the worst-case differences across all suppliers. The proof of Theorem~\ref{theorem:bound} relies on tracking flow differences across iterations under constraints \eqref{inflow} and \eqref{outflow}.

If we further assume that every consumer involved in the elimination of a supplier $s_i$ has not received any weight from any other suppliers until the iteration where $s_i$ is eliminated, we can claim the exactness of the approximation by Alg.~\ref{alg:nns} with respect to the optimal EMD as stated in Theorem \ref{theorem:exact}.

\begin{theorem}[\textbf{Exactness of EMD approximation by NNS-EMD}]\label{theorem:exact}
\setstretch{1.1}
    Let $\smash{M_{\widetilde{\mathbf{a}}^{(u)}_i[l]}}$ denote the accumulated consumed weight in the ${\widetilde{\mathbf{a}}^{(u)}_i[l]}$-th bin in $C$ immediately before the $u$-th iteration. Let $\smash{\widetilde{\mathbf{a}}^{(t_\Delta)}_i[k_i^{(t_\Delta)}]}$ be the index of the last consumer from $C$ that receives a positive flow from $s_i$ before $s_i$ is eliminated, with $1\!\leq \!\!k^{(t_\Delta)}_i \!\!\leq\!\! |\mathbf{a}^{(t_\Delta)}_i\!|$. Then for the $i$-th supplier from $S$ in Lemma~\ref{lemma}, if $l \in \{1, 2, \ldots, |\smash{\widetilde{\mathbf{a}}^{(u)}_i|}\}$ with $u \in \{t_1, t_2, \ldots, t_{\Delta-1}\}$ or $l \in \{1, 2, \ldots, k_i^{(u)}\!\!-1\}$ with $u=t_\Delta$, 
\begin{equation}\label{eqn:F*}
    F^{*}_{i, \widetilde{\mathbf{a}}^{(u)}_i[l]} = c_{\widetilde{\mathbf{a}}^{(u)}_i[l]}-M_{\widetilde{\mathbf{a}}^{(u)}_i[l]}.
\end{equation}
If we assume $M_{\widetilde{\mathbf{a}}^{(u)}_i[l]}\!\!=0$ holds, we have $\smash{F^{*}_{i, \widetilde{\mathbf{a}}^{(u)}_i[l]} \!= c_{\widetilde{\mathbf{a}}^{(u)}_i[l]}}$, and then for any feasible flows $\mathbf{F}$,
\begin{align}
    &\textstyle\sum_{j}F^*_{i,j}D_{i,j} - \sum_j F_{i,j}D_{i,j}\notag\\
    \leq\;& \textstyle D_{i, \widetilde{\mathbf{a}}^{(t_\Delta)}_i[k^{(t_\Delta)}_i]}\Big(\sum_{j}F_{i,j}^* - \sum_{j}F_{i,j}\Big)=0. \label{eqn:thm2} 
\end{align}
Therefore, $R = \sum_{i,j} F^{opt}_{i,j}D_{i,j}$.
\end{theorem}
The proof sketch for Theorem~\ref{theorem:exact} is as follows. Under the assumption that $M_{\widetilde{\mathbf{a}}^{(u)}_i[l]}\!\!=\!0$, the NNS operation in step~\ref{step:NNS} of Alg.~\ref{alg:nns} ensures that over the $\Delta$ iterations outlined in Lemma~\ref{lemma}, the distances between the $i$-th supplier in $S$ and its matched consumers are non-decreasing, which serves as the basis for the inequality in Eq.~\eqref{eqn:thm2} to hold. The equality to zero in Eq.~\eqref{eqn:thm2} follows directly from the feasibility of both $\mathbf{F}^*$ and $\mathbf{F}$, where $\sum_{j}F_{i,j} = s_i = \sum_{j}F^*_{i,j}$ per constraint \eqref{outflow}.

\subsection{Time Complexity Analysis}\label{sec:time}
Using the same notation, let the numbers of bins in $S$ and $C$ be $m$ and $n$, respectively; then the size of the distance matrix $\mathbf{D}$ is $m\times n$. Without loss of generality, we assume $n\geq m$. Theorem~\ref{thm:TC} presents the theoretical analysis of the upper bound on the time complexity (TC) for \alg, covering both the RP and the GP. 
\begin{theorem}[Time Complexity of \alg]\label{thm:TC}
\setstretch{1.1}
    Let the weights in $S$ and $C$ be initialized uniformly, i.e. $s_i = m^{-1}$ for $i \in \{1, 2, \ldots, m\}$ and $c_j = n^{-1}$ for $j \in \{1, 2, \ldots, n\}$. Assume that in each iteration, at least $\alpha m$ suppliers, where $\alpha \in (0,1]$, are paired with at least $n/m$ consumers. Then the total time complexity of \alg is upper bounded by $T_{\text{NNS}}+T_{\text{sorting}}$ for the GP, and $T_{\text{NNS}}$ for the RP, where
    \small
    \begin{align}
        T_{\text{NNS}}=& \frac{mn(2+\alpha)(1+\alpha)}{6\alpha};\label{eqn:TC_NNS}\\
        T_{\text{sorting}}=&\frac{2n\log n\!-\!n}{4\alpha}\!+\!\frac{6n\log n\!+\!3\alpha n\!-\!(1\!+\!\log \alpha n)}{12}\!+\!O(\frac{1}{n}).\label{eqn:TC_sorting}
    \end{align}
\end{theorem}
\normalsize
The proof of Theorem~\ref{thm:TC} is based on tracking the number of remaining suppliers and consumers over NNS iterations until convergence. The detailed proof is given in the Supplementary Materials.

In practical applications, histograms being compared are often of similar scale; i,e., $k$ is typically equal or close to 1. This ensures that the assumption that at least $\alpha m$ suppliers are paired with at least $n/m$ consumers can be easily satisfied and the results in Theorem~\ref{thm:TC} are widely applicable.

For GP, the dominant term between $T_{\text{NNS}}$ and $T_{\text{sorting}}$ depends on both $n/m$ and $\alpha$. When $n/m$ is small, meaning fewer consumers are paired per supplier, the total number of pairwise comparisons in NNS over all iterations increases, leading to $T_{\text{NNS}}$ (which scales as $O(n^2)$) being more dominant. Conversely, when $k$ is large, meaning more consumers are paired with a supplier, the burden on NNS decreases and $T_{\text{sorting}}$ for sorting becomes more dominant, which scales as $O(n\log n)$. Additionally, a smaller $\alpha$ slows convergence by limiting the number of suppliers involved in weights flow and updates per NNS iteration, further amplifying the impact of $T_{\text{NNS}}$. Since both $k$ and $\alpha$ are intrinsically determined by the underlying structures of the histograms rather than being tunable hyperparameters, the relative dominance of the two terms varies case by case.

For RP,  $T_{\text{sorting}}$ does not exist and then TC is solely determined by $T_{\text{NNS}}$. This makes RP computationally more efficient than GP in scenarios where sorting overhead is significant.\\
We also summarize the theoretical TC of various SOTA EMD approximation methods, along with the exact EMD\cite{flamary2021pot} in Table~\ref{tab:emd_complexities}. \alg improves over the TC of the exact EMD, which is $O(n^3\log n)$ to $T_{\text{NNS}}+T_{\text{sorting}}$, substantially reducing the computational burden for large datasets.

While some methods listed in Table~\ref{tab:emd_complexities} claim linear or near-linear TC, the theoretical results do not always translate into practical performance. Many of these methods rely on hyperparameter-related assumptions, require extensive hyperparameter tuning, or introduce significant pre-processing overhead not reflected in the TC results in Table~\ref{tab:emd_complexities}. 

For example, Diffusion\cite{tong2021diffusion} assumes that the diffusion operator $\mathbf{P}$ is sparse, containing at most $\Tilde{O}(n) = O(n\log^c n)$ nonzero entries. However, both $b$ and $c$ are implicitly determined by multiple hyperparameters, making it difficult to predict the actual computational cost.

\begin{table}[!b]
\centering
\caption{Summary of theoretical time complexities of SOTA EMD approximation algorithms.}
\renewcommand{\arraystretch}{1.1}
\footnotesize
\resizebox{0.48\textwidth}{!}{\begin{tabular}{c|c}
\toprule
\textbf{Algorithm} & \textbf{Time Complexity} \\
\midrule
Exact EMD \cite{flamary2021pot} & $O(n^3\log n)$ by linear programming \\ 
ACT \cite{atasu2019linear} & $O(n^2\log k + nk)$ \\
Diffusion \cite{tong2021diffusion} & $O(n\log^{2c}n + n\log^bn)$ \\
Flowtree \cite{tree} & $O(n(d+ \log(d\Phi)))$ \\
M3S \cite{chen2022computing} & No proof, seemingly linear only from the plot  \\
H-SWD \cite{nguyen2023hierarchical} & $O\left(H k d+H k L+H L\right)$ \\
\bottomrule
\end{tabular}}
\label{tab:emd_complexities}
\renewcommand{\arraystretch}{1}
\end{table}
{HSWD\cite{nguyen2023hierarchical} claims a theoretical TC independent of $n$ and is fully determined by the hyperparameters $(d, k, L, H)$. More importantly, it involves a ResNet-based feature extraction process that is not accounted for in its theoretical TC, which can significantly impacts the execution time in practice. For instance, the HSWD computation for two randomly selected $32\times32$ color images from CIFAR-10 following the original implementation with $d=8192, L=1000, k=70, H=1$ took 5.28 seconds, whereas the feature extraction alone required 8.92 seconds, which is not factored in the TC results. Finally, while Flowtree\cite{tree} shows an efficient theoretical TC of $O(n(d+ \log(d\Phi)))$, it exhibits the lowest accuracy among all evaluated methods in the experiments, limiting its practicality in applications requiring high-precision transport distance estimation.}

\subsection{Accelerating \alg on GPU} \label{sec:improvement}

We enhance \alg with vectorization on GPU, including parallelized NNS operation at the data point level and batch processing at the dataset level.

\textbf{Parallel NNS.} Based on Sec.~\ref{sec:time}, the bottleneck of \alg is the NNS operation with a quadratic time complexity. As discussed in prior works \cite{li2015brute,ravi2020pytorch3d}, the distance computations have no mutual dependencies. Therefore, we allocate the NNS operation for each consumer among $S$ to different GPU threads for parallel execution, which improves the execution time from quadratic to linear. 


\textbf{Batch Processing.} 
To further improve the computational efficiency of \alg on dataset levels, we use batch processing to reduce execution time. When computing \alg between $M$ supplier datasets and $N$ consumer datasets, we need to execute the algorithm $O(MN)$ times with a straightforward GPU implementation, which is inefficient for large datasets. By parallelizing computations across batches, the execution times can be reduced to $O(Bb^2)$, where $B$ is the number of batches and $b$ is the batch size (satisfying $bB = \max\{M, N\}$).

However, the zero-weight data points need to be eliminated in the weights update step as discussed in Sec.~\ref{sec:nns-emd}. Instead of removing them from $\mathbf{D}$ directly, 
we assign penalty terms to these points to exclude them from the subsequent NNS iterations. This approach allows us to keep the size of $\mathbf{D}$ consistent across each iteration, facilitating data unrolling for nested loops via vectorization.

\section{Experiments}
We conduct extensive experiments to evaluate \alg against an exact EMD algorithm, several SOTA approximate EMD methods, and a SOTA transformer model with $L_2$ distance. For the classification tasks examined in Sec.~\ref{sec:classification}, we perform both image and document classification on two public datasets for each data type.
We also benchmark \alg on image retrieval tasks in Sec.~\ref{sec:retrieval}. Besides,  we leverage the transport mapping generated from \alg to achieve color transfer between images in Sec.~\ref{sec:colortransfer}, and compare memory usage in Sec.~\ref{sec:memory}. Lastly, we investigate the sensitivity of \alg's performance in Sec.~\ref{sec:robustness} with respect to: 1) robustness to noise, 2) the impact of batch processing. 3) the choice of  ground distance ($L_1$ or $L_2$) for NNS operation.

\subsection{Experimental Setup}

\textbf{Datasets.}
Classification and retrieval tasks are the most common applications to evaluate exact EMD and approximate EMD algorithms in literature \cite{tong2021diffusion,tree,atasu2019linear, nguyen2023hierarchical}. Following the established experimental settings,
we use the MNIST \cite{lecun1998gradient} and CIFAR-10 \cite{krizhevsky2009learning} datasets for image classification, 20news \cite{lang1995newsweeder} and Amazon review \cite{mcauley2013hidden} datasets for document classification in Sec.~\ref{sec:classification}. For a more comprehensive comparison, we also evaluate the image retrieval application in Sec.~\ref{sec:retrieval} using the NUS-WIDE \cite{chua2009nus} and Paris-6k \cite{philbin2008lost} datasets in line with similarity retrieval research by \cite{zhu2016deep,xu2018iterative}. Additionally, in alignment with the M3S method in \cite{chen2022computing}, we employ the DOTmark dataset \cite{schrieber2016dotmark} for memory usage comparisons in Sec.~\ref{sec:memory}. These datasets are summarized in Table~\ref{tab:dataset} and more detailed descriptions are provided in the Supplementary Material.

{\textbf{Histogram Initialization.} \label{sec:his} We outline the steps for histogram construction -- defining bins, normalizing bin weights, and determining bin coordinates for calculating $\mathbf{D}$ -- for each data type used in the experiments. 1) \textit{grayscale images}: we use the spatial location-based histogram, where each pixel serves as a bin, its weight corresponds to the normalized pixel value, and its coordinate is the pixel's spatial coordinates; 2) \textit{RGB images}: we use a color intensity-based histogram, where each unique color serves as a bin, its weight is given by the normalized frequency of that color in the image, and its bin coordinate is the RGB values; 3) \textit{documents}: each unique word after text preprocessing (e.g., stop-word removal) forms a bin, its weight is the normalized term frequency (word count divided by the document length), and its bin coordinate is the vector generated from the GloVe word embeddings~\cite{pennington2014glove}.}

\begin{table}[t]
\centering
        \caption{Datasets description}\label{tab:dataset}
        \setlength{\tabcolsep}{2pt}
        \resizebox{0.46\textwidth}{!}{%
        \begin{tabular}{c|ccc}
        \toprule
        \textbf{Image datasets}& \textbf{\# of Images} & \textbf{\# of Queries}  & \textbf{Image Size}                     \\ \hline
         MNIST    &  60,000                &  1,000              &  {[}28, 28{]}                    \\
         CIFAR-10 &  50,000                &  1,000              &  {[}3, 32, 32{]}                  \\
         NUS-WIDE &  205,334               &  500              &  {[}3, 128, 128{]}                \\
         Paris-6k  & 6322              &  70              &  {[}3, 1024, 768{]}                  \\
         DOTmark  &  500                      &  500                   &  {[}32, 32{]} to {[}512, 512{]} \\ 
        \toprule
        \textbf{Text datasets}& \textbf{\# of Documents} & \textbf{\# of Queries}  & \textbf{Avg. \# of words}                     \\ \hline
         20news  &   11,314  &   1,000   &  115.9 \\
         Amazon review &   10,000  &   1,000   &  57.44 \\ 
        \bottomrule
        \end{tabular}}
\end{table}

\begin{table}[tb]
    \centering
    \renewcommand{\arraystretch}{0.75}
    \caption{Comparison of \alg and the exact EMD in terms of execution time and accuracy on image classification.}\label{tab:acceleration}
    \scriptsize
    \setlength{\tabcolsep}{10pt}
    \resizebox{0.46\textwidth}{!}{%
    \begin{tabular}{cccc}
    \toprule
    \multicolumn{1}{c}{\multirow{3}{*}{\textbf{Dataset}}} & \multicolumn{3}{c}{\textbf{Execution Time (s) (Accuracy)}} \\
    \cline{2-4}
    \addlinespace
    & Exact EMD & NNS-EMD & Speedup \\
    \midrule
    \multirow{2}{*}{MNIST} & \scriptsize 106.47 & \scriptsize \textbf{2.39} & \scriptsize44.55$\times$\\
    & \multicolumn{1}{c}{\scriptsize(99.12\%)} & \multicolumn{1}{c}{\scriptsize(97.45\%)} & \\
    \midrule
    \multirow{2}{*}{CIFAR-10} & \scriptsize1163.50 & \scriptsize \textbf{8.57} & \scriptsize135.77$\times$ \\
     & \multicolumn{1}{c}{\scriptsize(85.34\%)} & \multicolumn{1}{c}{\scriptsize(82.12\%)} & \\
    \bottomrule
    \end{tabular}%
    } 
\end{table}

\textbf{Algorithms and Implementations.}
\alg is implemented with the greedy protocol (GP) by default, with comparisons to the random protocol (RP) in image classification (Table~\ref{tab:class_res}) and retrieval tasks (Table~\ref{tab:class_res}). For the ground distance in NNS operation, we use $L_2$ distance for higher accuracy (based on the results from Table~\ref{tab:relative_error}). 
We implement the exact EMD \cite{flamary2021pot} and various SOTA approximate EMD algorithms, including Flowtree \cite{tree}, Diffusion EMD\cite{tong2021diffusion}, ACT \cite{atasu2019linear}, M3S \cite{chen2022computing}, and H-SWD \cite{nguyen2023hierarchical}. Among them, the exact EMD, Flowtree, and Diffusion methods are challenging to be accelerated using GPU due to the inherent data dependencies in each iteration. Therefore, they are implemented on CPU following the original implementations in their papers, while ACT, M3S, and H-SWD are parallelized on GPU to ensure a fair comparison with \alg.

Moreover, to demonstrate the need for efficient and accurate EMD approximation, in the image classification and retrieval tasks, we also contrast the aforementioned EMD-based methods with a SOTA deep learning model combined with a simpler distance function than EMD. Specifically, we fine-tune the Vision Transformer (ViT) model in its basic version \cite{dosovitskiy2020image} from a pre-trained model. We then extract image features through the ViT model and use the $L_2$ norm as the distance metric. The batch size and hyper-parameter settings follow those specified in the original work. 

\begin{table*}[bt]
\Large
\centering
\caption{Comparison of different approximate EMD algorithms for image classification and image retrieval.\label{tab:class_res}}
\renewcommand{\arraystretch}{1.02}
\resizebox{0.98\textwidth}{!}{
\begin{tabular}{c|ccccc|ccccc|ccc|ccc}
\toprule
& \multicolumn{10}{c|}{\textbf{Image classification}}& \multicolumn{6}{c}{\textbf{Image retrieval}}\\\cline{2-17}
& \multicolumn{5}{c|}{\textbf{MNIST}} & \multicolumn{5}{c|}{\textbf{CIFAR-10}} & \multicolumn{3}{c|}{\textbf{NUS-WIDE}} & \multicolumn{3}{c}{\textbf{Paris-6k}}\\
{\multirow{-3}{*}{\textbf{Algorithm}}} & Accuracy$\uparrow$ & Precision$\uparrow$ & Recall$\uparrow$ & F1$\uparrow$ & Time (s)$\downarrow$ & Accuracy & Precision & Recall & F1 & Time (s) & Recall$\uparrow$ & MAP$\uparrow$ & Time (min)$\downarrow$& Recall & MAP & Time (s)\\ \hline
ACT \cite{atasu2019linear} &  91.05\%$^\dagger$ &  0.903 &  0.901 & 0.902 &  20.87 &  71.68\% &  0.705 &  0.701 &  0.702 & 84.62 &  0.727 & 0.630 &  36.44 &  0.605 &  0.542 & 314.23 \\
M3S \cite{chen2022computing} &  95.87\% & 0.951 &  0.948 & 0.949 &  16.93 &  76.98\% &  0.768 &    0.760 &  0.764 &  78.50 &  0.754 & 0.646 &  29.16  &    0.625 &  0.583 &  247.36 \\
Flowtree \cite{tree} &  68.45\% &  0.663 &  0.666 &  0.665 &  0.94 &   46.57\% &  0.459 &  0.450 & 0.455 & 5.42 &  0.519 &  0.457 &  6.14 &  0.382 & 0.417 & 22.72 \\
Diffusion \cite{tong2021diffusion} &  83.21\% &  0.830 &   0.830 &  0.830 &   1.34 &  64.21\% &  0.640 &  0.639 & 0.640 &  8.98 &   0.646 &  0.571 &   13.51 &  0.459 & 0.438 &  27.18 \\
H-SWD \cite{nguyen2023hierarchical} &  86.78\% &  0.861 &  0.862 &  0.861 &  2.80 &  70.24\% &  0.699 &   0.699 & 0.699 &  13.57 &  0.681 &  0.613 &  7.52  &   0.531 & 0.554 &  10.43\\ \midrule
NNS-EMD$^R$ & 95.19\% & 0.950 & 0.951 & 0.950 & 1.57 & 76.42\% & 0.759 & 0.761 & 0.760 & 6.13 &  0.701 & 0.589 & 5.47  & 0.565 & 0.531 & 47.46\\
NNS-EMD$^G$ & \textbf{97.45\%} & \textbf{0.974} & \textbf{0.975} & \textbf{0.972} & 2.39 & \textbf{82.12\%} & \textbf{0.816} & \textbf{0.815} & \textbf{0.816} & 8.57 & 0.809 & 0.696 & 8.94 & \textbf{0.661} & \textbf{0.647} & 95.73\\\midrule

ViT+$L_2$ (0-epoch) & 18.57\% & 0.167 & 0.179 & 0.161 & \textbf{0.42}  & 16.43\% & 0.168 & 0.142  & 0.138 & \textbf{1.02} & 0.347 & 0.204 & \textbf{0.45} & 0.085  & 0.003  & \textbf{1.07} \\

ViT+$L_2$ (1-epoch) & 31.42\% & 0.286 & 0.321 & 0.303 & 1.67 & 28.37\% & 0.307 & 0.282 & 0.247 & 3.44 & 0.683 & 0.534 & 5.07 & 0.139 & 0.157 & 4.78\\
ViT+$L_2$ (3-epoch) & 51.87\% & 0.520 & 0.523 & 0.522 & 5.43 & 36.52\% & 0.377 & 0.361 & 0.345 & 9.72 & 0.741 & 0.603 & 15.42 & 0.328 & 0.411 & 13.14\\
ViT+$L_2$ (5-epoch) & 72.13\% & 0.723 & 0.721 & 0.722 & 8.44 & 40.63\% & 0.407 & 0.402 & 0.395 & 16.33 & 0.762 & 0.743 &25.78 & 0.567 & 0.519 & 21.04 \\
ViT+$L_2$ (10-epoch) & 96.42\% & 0.961 & 0.958 & 0.960 & 17.17 & 79.17\% & 0.790 & 0.788 & 0.784 & 32.67 & \textbf{0.821} & \textbf{0.835} &51.66 & 0.591 & 0.545 & 42.08 \\
\bottomrule
\multicolumn{11}{p{1.0\linewidth}}{\large NNS-EMD$^R$: \alg with random protocol; NNS-EMD$^G$: \alg with greedy protocol.}\\
\multicolumn{17}{p{2\textwidth}}{\large
$^\dagger$: Note that the $91.05\%$ accuracy here is lower than the $97.42\%$ reported in the original ACT paper \cite{atasu2019linear}. First, since the code for ACT is not publicly available, our implementation was based on the replication released by \cite{tree} (with 90\% reported in \cite{tree}). Second, the query images used in \cite{atasu2019linear} were the first $6k$ training images –- a subset of the $60k$ training set -– whereas our query images are the $1k$ test images. 
We did not follow the experimental settings in the ACT paper because we believe the query images should be different from the training images in order to make the experimental results more convincing. When using the same experimental configuration as in the original ACT paper, our code achieves $97.52\%$ accuracy and takes 120.42s, compared to the $97.76\%$ accuracy and $\sim$75s reported in~\cite{atasu2019linear}.} 
\end{tabular}}
\renewcommand{\arraystretch}{1}
\end{table*}

\begin{table}[t]
\centering
\small
\caption{Comparison on document classification datasets. \label{tab:doc_class}}
\renewcommand{\arraystretch}{1}
\resizebox{0.48\textwidth}{!}{
\begin{tabular}{c|cc|cc}
\toprule
& \multicolumn{2}{c|}{\textbf{20news}} & \multicolumn{2}{c}{\textbf{Amazon review}} \\
\multicolumn{1}{c|}{\multirow{-2}{*}{\textbf{Algorithm}}} & Accuracy & \begin{tabular}[c]{@{}c@{}}Time (s)\end{tabular} & Accuracy & \begin{tabular}[c]{@{}c@{}}Time (s)\end{tabular} \\ \hline
ACT \cite{atasu2019linear} & 87.42\%$^\dagger$ & 2.23 & 78.53\% & 0.74 \\

Flowtree \cite{tree} & 75.18\% & \textbf{0.24} & 71.69\% &\textbf{0.09} \\
M3S \cite{chen2022computing} & 89.55\% & 4.93 & 86.43\% & 1.58 \\\hline
NNS-EMD$^G$ & \textbf{91.88\%} & {1.89} & \textbf{93.24\%} & {0.55} \\
\bottomrule
\multicolumn{5}{p{0.48\textwidth}}{\scriptsize $^\dagger$: Note that the 87.42\% accuracy here is different from the 83\% reported in the original ACT paper \cite{atasu2019linear}, since we use different word embeddings. In \cite{atasu2019linear}, Word2Vec \cite{mikolov2013efficient} was used to convert words into 300-dimensional real values, while we use GloVe \cite{pennington2014glove} to convert words into 50-dimensional real values, which is faster than the 300-dimensional data. If we use the same configuration as in the original ACT paper, our implemented ACT achieves 84.12\% accuracy and takes 83.42s, compared to the 83\% accuracy and $\sim$90s reported in \cite{atasu2019linear}.}
\end{tabular}%
}\vspace{-12pt}
\end{table}


{In our experiment, we employ the following classification approaches: (1) For the exact and approximate EMD-based method, we leverage a traditional feature-based approach (i.e., manually constructing the histograms and distance matrices) to compute the EMD distances. We then use the $k$-NN method ($k=3$) to identify the majority label based on the calculated EMD distances. (2) For the deep learning method, we use the ViT model to learn the representations or features of the image and then predict the label.
}

\textbf{Metrics.}
For image classification, we adopt accuracy, precision, recall, and F1-score as performance metrics \cite{tripathi2021analysis}. For the document classification, we adopt the accuracy as the metric \cite{tree}. For image retrieval, we use two widely-adopted metrics \cite{zheng2015scalable}: recall and mean average precision (MAP). Execution time is measured by the average execution time for processing a single query on the datasets. Detailed metric explanations are given in the Supplementary Material. 

\textbf{Platforms.}
We run the aforementioned CPU-based implementations on an Intel i9-12900K CPU with 128 GB of DDR4 memory, and the GPU-based implementations on an NVIDIA 3080Ti GPU with 12 GB of memory.

\subsection{Classification of Images and Documents}\label{sec:classification}

\textbf{Image Classification.} In this task, we first compare our \alg with the exact EMD. Table~\ref{tab:acceleration} shows the execution time and accuracy results of exact EMD and \alg. \alg achieves 44.55$\times$ and 135.77$\times$ speedup over an exact EMD realization on the MNIST and CIFAR-10 datasets, respectively, while incurring an accuracy loss of only 1.67\% and 3.22\%. 

\alg is also benchmarked against SOTA approximate EMD algorithms. In Table \ref{tab:class_res}, our experimental results show that \alg is superior to all other approximate EMD methods in every metric except when compared to the execution time of the Flowtree and Diffusion methods. Specifically, \alg with GP not only achieves $\sim$1.6\% and $\sim$5\% higher accuracy but also attains speedup of 7.08$\times$ and 9.16$\times$ than the next-best algorithm (M3S) on MNIST and CIFAR-10, respectively. 
Note that, although \alg with GP is 2.54$\times$ and 1.58$\times$ slower than the Flowtree solution, it is also 29\% and 36\% more accurate than the Flowtree method on MNIST and CIFAR-10, respectively. 

{Comparing the RP and GP used in \alg, we observe that GP consistently outperforms RP in terms of accuracy but takes a longer time, with the extent of the trade-off varying across tasks. Specifically, on the MNIST dataset, Greedy improves accuracy by $\sim\! 2\%$ but incurs $\sim\!52.2\%$ additional computation time over RP. For the CIFAR-10 dataset, GP improves accuracy by $\sim \!5.7\%$ but requires $\sim \!39.8\%$ additional computation time over RP. Therefore, RP is a better choice if the running time is a critical concern, otherwise GP is preferred since it generally provides higher accuracy.}

We further compare the performance of \alg with a more recent image classification method: a SOTA Transformer network with $L_2$ distance \cite{dosovitskiy2020image} (i.e., ViT+$L_2$ in Table~\ref{tab:class_res}). {The reported time for ViT includes fine-tuning and inference time but excludes the original pre-training time of the ViT model. As shown in the row of ``ViT+$L_2$ (0-epoch)'' in Table~\ref{tab:class_res}, though the running time (i.e., inference-only time) is consistently the shortest, the accuracy is consistently being the lowest across all experiments, e.g. achieving only $18.57\%$ accuracy on MNIST. Thus, to ensure a fair comparison, we explored and reported different fine-tuning epochs (1, 3, 5, and 10) to provide a comprehensive view of the total time required for ViT to reach a competitive accuracy level.} 

For the MNIST dataset, we observe that ViT+$L_2$ trained for 10 epochs shows quite similar accuracy as \alg with Greedy Protocol (GP) (96.42\% compared to 97.45\%). However, the time required by ViT+$L_2$ is much larger than that of \alg with GP (17.17s compared to 2.39s). On the CIFAR-10 dataset, ViT+$L_2$ gives lower accuracy and requires more time compared to \alg with GP. 
While we acknowledge that the pre-trained ViT+$L_2$ can achieve a higher accuracy \cite{dosovitskiy2020image}, as shown by current leaderboard results\footnote{https://paperswithcode.com/sota/image-classification-on-cifar-10}, such improvement typically requires more training epochs and uses additional large-scale datasets (e.g., ImageNet \cite{krizhevsky2012imagenet}), hence may significantly increase latency. For instance, as stated in the original paper \cite{dosovitskiy2020image}, the ViT model could be trained using a standard cloud TPUv3 with 8 cores in approximately 30 days. In contrast, our proposed algorithm does not require dataset training and careful parameter tuning (such as the learning rate in the optimizer for ViT).

\textbf{Document Classification.} To investigate the performance and scalability of \alg across different data types, we further conduct document classification experiments on two text classification datasets. The results in Table~\ref{tab:doc_class} show that NNS-EMD achieves the highest accuracy on both datasets. Similar to the image classification results in Table~\ref{tab:class_res}, Flowtree also exhibits the fastest runtime, at the cost of significantly reduced accuracy.
In contrast, M3S has the second highest accuracy (still notably lower than \alg) but requires more than twice the runtime of \alg. 

In summary, \alg not only offers a better balance between accuracy and computational efficiency, but also demonstrates robust performance across CV and NLP applications and different datasets, holding promise for broader applicability in diverse domains.

    
    

\subsection{Image Retrieval}\label{sec:retrieval}

We also conduct experiments on the image retrieval task by benchmarking \alg against other EMD approximations and ViT+$L_2$. To comprehensively evaluate the scalability and adaptability of these methods, we choose two datasets: the NUS-WIDE dataset \cite{chua2009nus} including a large collection of images, and the Paris-6k dataset \cite{philbin2008lost} consisting of a smaller set of high-resolution color images. 

The experimental results are presented in Table~\ref{tab:class_res}. Observations indicate that \alg achieves the highest recall and MAP among all the compared approximate EMD methods. In the NUS-WIDE dataset, \alg with RP is faster than the Flowtree method. This efficiency is attributed to \alg's utilization of batch processing, particularly beneficial for large datasets like NUS-WIDE. For the Paris-6k dataset, although the H-SWD method leads in speed by projecting high-resolution images into one-dimensional data, its recall and MAP are significantly worse than \alg. 

Comparing \alg and ViT+$L_2$, we observe that \alg offers a better balance between accuracy and speed across various datasets. On the NUS-WIDE dataset, which contains over 200k images of size \([3,128, 128]\), though ViT+$L_2$ (10-epochs) exhibits slightly higher recall and MAP, it requires nearly 6 times the runtime that \alg takes. In contrast, on the Paris-6k dataset with large-size images of size \([3,1024,768]\), NNS-EMD significantly outperforms ViT+\(L_2\) (10-epochs) in accuracy. Although ViT+\(L_2\) (10-epochs) shows a faster execution time, this speed advantage largely stems from resizing the original high-resolution images to smaller standardized dimensions. In DL models, resizing input images to unified dimensions, such as \([3,224,224]\) in this case, is typically required by most methods. However, \alg does not need any resizing operation and directly processes images in their original size, leading to a slower but much more accurate performance relative to ViT+\(L_2\). 

Overall, \alg demonstrates more consistent and efficient performance across various scales of datasets and tasks, offering a compelling alternative to both traditional EMD approximations and modern deep learning approaches. Additionally, we note that as the image size increases, \alg with RP achieves greater speedup than with GP. Specifically, RP achieves $1.39\times$, $1.63\times$, and $2.02\times$ speedup compared to GP on the CIFAR-10, NUS-WIDE, and Paris-6k datasets, respectively. This is because the additional ranking operation in GP becomes more time-consuming as the image size gets bigger.

\begin{figure*}[bt]
    \centering
    \includegraphics [ scale=0.49] {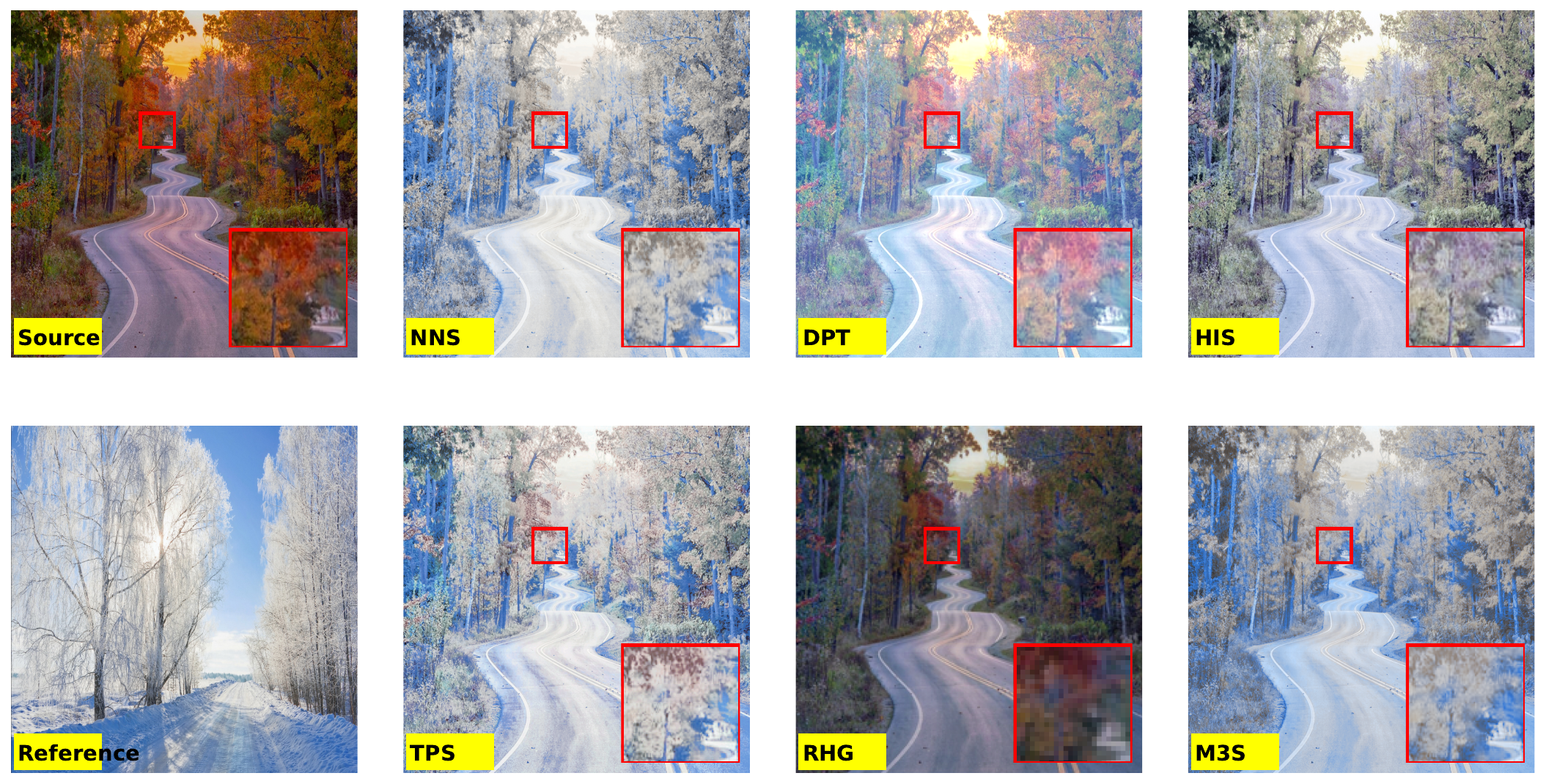}
    
    \caption{Qualitative comparison of color transfer results. The top-left image is the source image, and the bottom-left image is the reference image. The remaining images show color transfers produced by NNS (ours), DPT, HIS, TPS, RHG, and M3S. Red boxes indicate zoom-in regions for detailed inspection.}
    \label{fig:color_transfer_new}
\end{figure*}

\subsection{Color Transfer between Images} \label{sec:colortransfer}

Color transfer task is gaining interest in computer vision and is widely applied in real-world applications, such as photo editing and underwater imaging~\cite{liu2022overview,reinhard2001color,li2018emerging,yang2022underwater}. 
The color transfer process involves recoloring a \textit{source} image using a transport mapping from a \textit{reference} image (supplier) to this source image (consumer). The flow matrix $\mathbf{F}^{opt}$ optimized by the exact EMD has been shown as an effective approach to identifying an optimal transport mapping ~\cite{ferradans2014regularized,blondel2018smooth,chen2022computing} for color transferring. However, most SOTA approximate EMD algorithms \cite{atasu2019linear, tree, nguyen2023hierarchical, tong2021diffusion} merely focus on estimating the value of EMD without computing the associated transportation map. In comparison, \alg can generate the transport mapping $\mathbf{F}^*$ for color transferring between images.

Following the image pre-processing steps in \cite{chen2022computing,blondel2018smooth}, we apply $k$-means clustering to reduce the reference image to $m$ distinct pixel vectors in RGB space. By adopting the color intensity-based histogram formulation introduced in Sec. \ref{sec:emd_bg}, there are $m$ bins for this color image. The weight of each bin is the normalized frequency of its pixel values and the coordinate is its corresponding RGB vector. After performing the same formulation on the source image with $n$ bins, \alg can be employed to output a transportation map $\mathbf{F}^*$.


{
To evaluate \alg’s color transfer performance both qualitatively and quantitatively, we use the seasonal dataset following \cite{chen2022computing}. This dataset contains 30 RGB images of $1024\times1024$, where each image can serve as either a source or a reference. As a result, we conduct 870 experiments in total (30 source images $\times$ 29 possible reference images).}

{
We compare \alg against 5 color transfer approaches, including both traditional and learnable methods, using 8 evaluation metrics, following the framework in~. The traditional methods are Tps Color Transfer (TPS)~\cite{grogan2019l2} and M3S~\cite{chen2022computing}, while the learnable methods include Deep Photo Style Transfer (DPT)~\cite{luan2017deep}, Histogram Analogy (HIS)~\cite{lee2020deep}, and Re-HistoGAN (RHG)~\cite{afifi2021histogan}. The 8 evaluation metrics -- PSNR~\cite{huynh2008scope}, MS-SSIM~\cite{wang2003multiscale}, FSIM~\cite{zhang2011fsim},  BA~\cite{goude2021example}, HI~\cite{potechius2021color}, Corr~\cite{huang2011automatic}, BRISQUE~\cite{mittal2012no}, and NIQE~\cite{mittal2012making} -- are selected from the 20 metrics in \cite{10.1145/3626495.3626509} to cover 3 different aspects: structural similarity, color fidelity, and perceptual quality. Detailed descriptions of each method and metric are provided in the Supplementary Material.}


{
We present the experimental results, demonstrating how \alg compares against benchmark methods both visually (qualitatively) and via objective metrics (quantitatively). Fig.~\ref{fig:color_transfer_new} shows a qualitative comparison of color transfer outputs across six methods: \alg, DPT, HIS, TPS, RHG, and M3S. Overall, \alg achieves a balanced adaptation of color, effectively incorporating the bluish tones of the reference while preserving key structural details, especially around the winding road and the surrounding foliage. In contrast, HIS and TPS maintain structural fidelity but exhibit some degree of color inconsistency in the tree regions. DPT and RHG produce vibrant recolorings, albeit at the expense of occasional over-saturation or block artifacts, as highlighted in the zoomed-in views. M3S yields generally coherent results, though it occasionally appears desaturated in certain portions of the foliage. These observations suggest that \alg provides a more visually consistent color mapping than many existing methods, while effectively preserving structural integrity and mitigating artifacts. More qualitative comparisons can be found in the Supplementary Material.}

{
As shown in Table~\ref{tab:quantitative}, NNS-EMD achieves the best performance in PSNR and MS-SSIM and ranks 2nd in FSIM, demonstrating strong structural similarity preservation. Also, \alg outperforms all other methods across all three color fidelity metrics and ranks 3rd place in two perceptual quality metrics, highlighting \alg's superior performance in particularly effective in aligning color distributions between the source and reference images.
Furthermore, \alg achieves the lowest average execution time (4.572 seconds) with the smallest standard deviation (0.236 seconds), demonstrating both efficiency and high stability. These combined advantages make NNS-EMD a robust solution for color transfer tasks.}

\begin{figure*}[!t]
\begin{minipage}{0.62\textwidth}
\captionsetup{type=table}
\centering
\caption{Mean (SD) of quantitative evaluation for color transfer. \label{tab:quantitative}}
\setlength{\tabcolsep}{2pt}
\renewcommand{\arraystretch}{1.23}
\normalsize
\resizebox{\textwidth}{!}{
\begin{tabular}{c|c c c|c c c|c c|c}
\toprule
\multicolumn{1}{c|}{\multirow{2}{*}{\textbf{Algorithm}}} 
& \multicolumn{3}{c|}{\textbf{Structure Similarity}} 
& \multicolumn{3}{c|}{\textbf{Color Fidelity}} 
& \multicolumn{2}{c|}{\textbf{Perceptual Quality}} 
& \multicolumn{1}{c}{\textbf{Execution}} \\

& PSNR$\uparrow$ & MS-SSIM$\uparrow$ & FSIM$\uparrow$
& BA$\downarrow$ & HI$\uparrow$ & Corr$\uparrow$ 
& BRISQUE$\downarrow$ & NIQE$\downarrow$ 
& \textbf{Time (sec)}$\downarrow$ \\
\hline
\multirow{2}{*}{NNS-EMD} 
& \textbf{11.902} & \textbf{0.215} & 0.569
& \textbf{0.024} & \textbf{0.963} & \textbf{0.237}
& 17.531 & 3.220
& \textbf{4.572} \\ 
& (1.711) & (0.066) & (0.041)
& (0.017) & (0.051) & (0.124)
& (12.037) & (1.147)
& (0.236) \\
\hline
\multirow{2}{*}{M3S~\cite{chen2022computing}} 
& 11.033 & 0.201 & 0.556
& 0.067 & 0.609 & 0.148
& 16.216 & 3.315
&  6.583\\ 
& (1.633) & (0.068) & (0.039)
& (0.073) & (0.072) & (0.110)
& (12.921) & (1.304)
& (0.318) \\

\hline
\multirow{2}{*}{HIS~\cite{lee2020deep}} 
& 10.545 & 0.170 & 0.560
& 0.059 & 0.769 & 0.042
& \textbf{15.392} & 3.185
&  5.424\\ 
& (1.474) & (0.071) & (0.040)
& (0.028) & (0.049) & (0.130)
& (12.945) & (1.163)
& (0.273) \\

\hline
\multirow{2}{*}{DPT~\cite{luan2017deep}} 
& 10.863 & 0.163 & \textbf{0.573}
& 0.174 & 0.877 & 0.042
& 23.964 & 3.412
& 314.275 \\ 
& (1.756) & (0.062) & (0.040)
& (0.013) & (0.037) & (0.122)
& (11.896) & (1.084)
& (18.921) \\

\hline
\multirow{2}{*}{RHG~\cite{afifi2021histogan}} 
& 9.242 & 0.188 & 0.554
& 0.200 & 0.577 & 0.052
& 50.614 & 8.089
& 8.798 \\ 
& (1.459) & (0.060) & (0.041)
& (0.140) & (0.133) & (0.122)
& (7.152) & (0.782)
& (0.632) \\

\hline
\multirow{2}{*}{TPS~\cite{grogan2019l2}} 
& 10.305 & 0.171 & 0.560
& 0.048 & 0.780 & 0.036
& 16.695 & \textbf{3.168}
& 34.291 \\ 
& (1.521) & (0.067) & (0.041)
& (0.039) & (0.079) & (0.120)
& (12.252) & (1.135)
& (4.163) \\
\bottomrule
\end{tabular}}
\end{minipage}
\hspace{-1.5in}
\begin{minipage}{0.77\textwidth} 
    \centering
    \vspace{15pt}
    \centerline{\includegraphics[width=0.44 \textwidth] {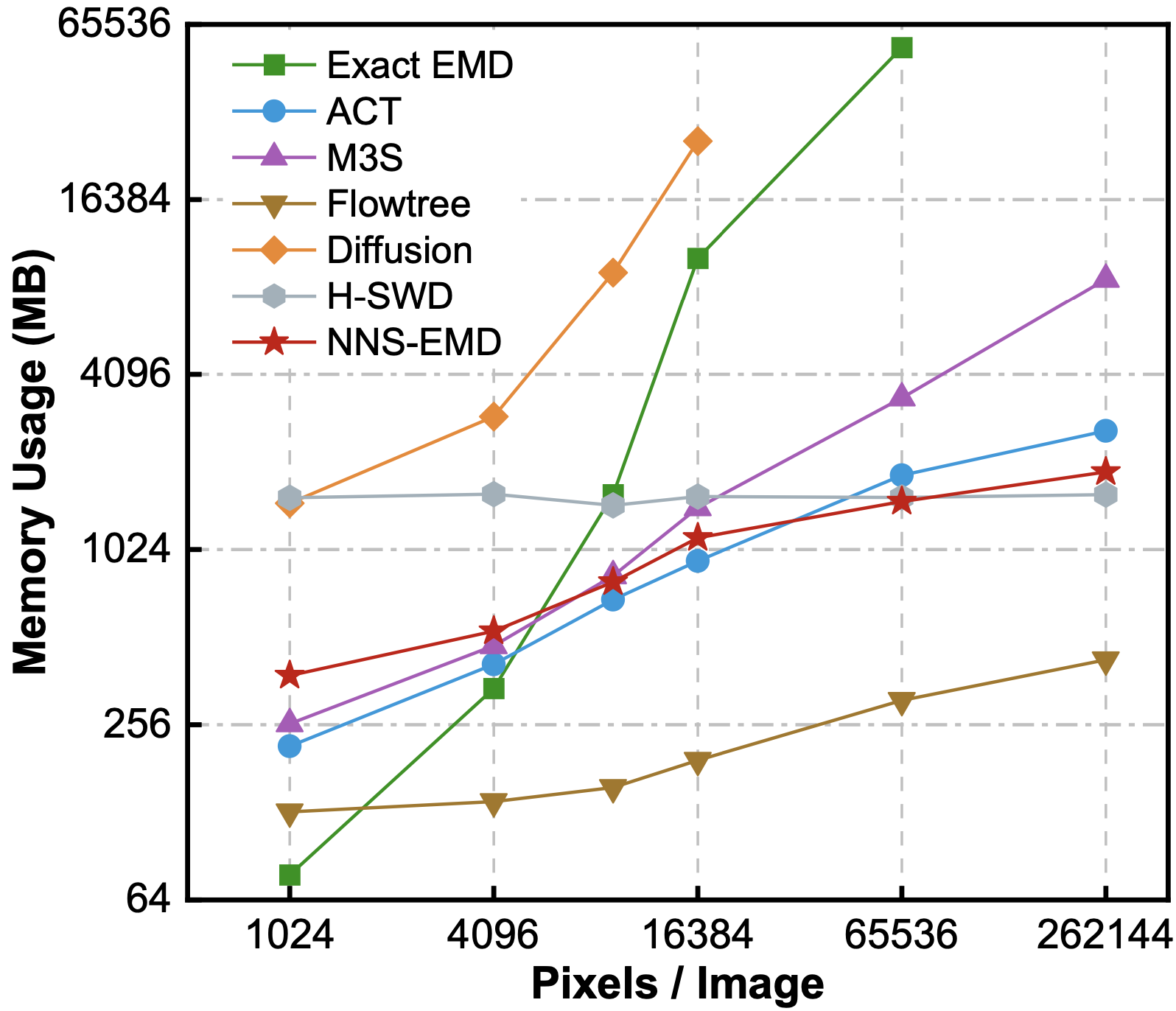}}
    \captionsetup{width=0.4\linewidth}
    \caption{Memory usage comparison of various approximate EMD algorithms.}
    \label{fig:memory_usage}
\end{minipage}
\end{figure*}\vspace{-6pt}

\subsection{Memory Usage Evaluation}\label{sec:memory}

In addition to the accuracy and performance metrics, we also evaluate the memory usage by \alg, the exact EMD solution, and other SOTA approximate EMD algorithms. We conduct experiments on the DOTmark dataset following M3S \cite{chen2022computing} with image sizes varying from 32$\times$32 to 512$\times$512 (i.e., from 1024 to 262,144 pixels). We employ the psutil library\footnote{psutil library: \url{https://github.com/giampaolo/psutil}} to measure the memory usage of CPU-based implementations, and Nvidia-smi for GPU-based implementations.  

As illustrated in Fig.~\ref{fig:memory_usage}, the \alg, M3S, and ACT methods exhibit a linear growth in memory requirements, with \alg and ACT having similar slopes, while M3S shows a steeper slope. Flowtree approximation demonstrates a sub-linear increase in memory usage.
The H-SWD approximation shows near-constant memory usage relative to the number of pixels, as its memory usage depends only on the number of projections. The diffusion and exact EMD approaches demonstrate a cubic increase in memory requirement, and exhaust available memory after image sizes of $256\times256$ and $512\times512$.
    

\subsection{Additional Considerations on \alg }

In this section, we explore additional aspects of \alg's performance. We first evaluate its robustness to various types of noise, then investigate how batch processing impacts the computational efficiency, and finally compare the accuracy of $L_1$ or $L_2$ distance as ground distance in the NNS operation. These investigations provide deeper insights into \alg's behavior and practical considerations for its application.

\textbf{Robustness to Noise.}\label{sec:robustness} Since the NNS operation within \alg may be influenced by nearby points, the presence of noisy data can potentially affect its accuracy in approximating EMD. To evaluate \alg's robustness to different types of noise, we employ the CIFAR-10-C dataset, a corrupted version of the CIFAR-10 dataset. Introduced as part of a robustness benchmark in \cite{hendrycks2019robustness}, the CIFAR-10-C dataset includes 15 types of corruptions (from 4 broad categories: noise, blur, weather, and digital) applied to test images, with each type featuring 5 levels of severity. Because the noise category in CIFAR-10-C consists of totally 3 types of noise, our experiments focus on these three noise types to assess the algorithms' resilience to noise-related corruptions. These three types of noise are:
1) Gaussian noise, which commonly appears in low-light conditions; 2) shot noise, also called Poisson noise, which is electronic noise caused by the discrete nature of light itself; 3) impulse noise, also known as salt-and-pepper noise, which is characterized by randomly-occurred white and black pixels in an image. Impulse noise can also be viewed as adding outliers, as detailed in \cite{mitra2010robust}.

Furthermore, 
since the EMD values are inherently dataset-specific, intentionally introducing noise can alter the ``target" EMD that approximation algorithms aim to estimate. By employing the exact EMD as a baseline and comparing it with each approximate EMD method at identical levels of perturbation, we can effectively evaluate how sensitive these approximate EMD algorithms are to noise. Here, we only compare \alg to the M3S method since M3S consistently outperforms the other SOTA approximate EMD algorithms in terms of accuracy, ranking second only to our \alg in the previous evaluations as shown in Tables~\ref{tab:class_res} and \ref{tab:doc_class}.

We use two corruption robustness metrics proposed in \cite{hendrycks2019robustness} for evaluating the aforementioned algorithms' robustness to each type of noise:
\begin{align}
    \text{CE}_c^f &= \frac{\sum_{s=1}^{5} E_{s, c}^{f}}{\sum_{s=1}^{5} E_{s,c}^{\text{ exact EMD}}},\label{eqn:CE}\\
    \text{Relative CE}_c^f &= \frac{\sum_{s=1}^{5} E_{s, c}^{f} - E_{\text{clean}}^{f}}{\sum_{s=1}^{5} E_{s,c}^{\text{ exact EMD}} - E_{\text{clean}}^{\text{ exact EMD}}}.\label{eqn:RCE}
\end{align}

\noindent
The Corruption Error (CE) in Eq.~\eqref{eqn:CE} for a method $f$ on a noise type $c$ is calculated as follows:  first, we record the top-1 classification error for algorithm $f$ on the ``clean'' CIFAR-10, denoted as $E^{f}_{\text{clean}}$, with $f \in$ \{exact EMD, M3S, NNS-EMD\}. Then, we test every method on each noise type $c$ at each level of severity $s$ ($1\leq s\leq 5$)\footnote{Some detailed information for the 5 levels of severity: For Gaussian noise, we use a mean of 0 and different standard deviations from $[0.04, 0.06, 0.08, 0.09, 0.10]$. For shot noise, we set the intensity parameter to $[500, 250, 100, 75, 50]$. For impulse noise, we set the noise density to $[0.01, 0.02, 0.03, 0.05, 0.07]$.} using CIFAR-10-C, and record the top-1 error as $E^{f}_{s,c}$. Lastly, the aggregated performance is adjusted for various severities via dividing by the baseline's errors. The Relative Corruption Error in Eq.~\eqref{eqn:RCE} is designed to account for a method's performance degradation under noise relative to its error on clean data. It helps identify methods that may have larger CE but degrade more gracefully in the presence of noise, providing a more comprehensive view of robustness.




Table~\ref{tab:robustness_metrics} demonstrates that NNS-EMD is more robust than M3S across all the three noise types as NNS-EMD shows lower CE and Relative CE, indicating its better resilience to noise. Specifically, under Gaussian noise, NNS-EMD has a CE of  119.08\% and a Relative CE of 120.99\%, both of which are significantly lower than those of M3S, which are 129.49\% and 129.73\% respectively. For impulse noise, \alg's CE is 104.29\% and its Relative CE is 104.59\%, indicating that \alg is more robust to outliers than M3S, whose CE is 138.75\% and Relative CE is 138.94\%. Furthermore, \alg shows the highest robustness against impulse noise, followed by shot noise, with Gaussian noise posing the greatest challenge.
This pattern suggests that NNS-EMD is more robust to localized and discrete perturbations (characteristic of impulse and shot noise) compared to the continuous corruptions of Gaussian noise. Overall, NNS-EMD exhibits substantial improvement in robustness over M3S across all the tested noise types.

\begin{table}[t]
    \centering
    \scriptsize
    \setlength{\tabcolsep}{3pt}
    \caption{Comparison of robustness in different noise types (Gaussian, Shot, and Impulse noise). Higher percentages indicate worse robustness to noise.}
    \label{tab:robustness_metrics}
    \renewcommand{\arraystretch}{1.2}
    \resizebox{\linewidth}{!}{%
        \begin{tabular}{c|ccc|ccc}
            \toprule
            \multirow{2}*{\textbf{Method}} & \multicolumn{3}{c|}{\textbf{CE (\%)$\downarrow$}} & \multicolumn{3}{c}{\textbf{Relative CE (\%)$\downarrow$}} \\
            & \textbf{Gauss.} & \textbf{Shot } & \textbf{Impulse} & \textbf{Gauss.} & \textbf{Shot} & \textbf{Impulse} \\
            \midrule
            Exact EMD\cite{flamary2021pot} & 100 & 100 & 100 & 100 & 100 & 100 \\
            NNS-EMD & 119.08 & 111.03 & 104.29 & 120.99 & 112.76 & 104.59\\
            M3S\cite{chen2022computing} & 129.49 & 141.32 & 138.75 & 129.73 & 142.12 & 138.94\\
            \bottomrule
        \end{tabular}
    }
\end{table}

\textbf{Batch Processing on GPU.}\label{sec:vectorization} We further evaluate the performance improvement of \alg obtained when using batch processing (as described in Sec.~\ref{sec:improvement}), especially on large datasets. We compare the execution time of \alg with or without batch processing across dataset sizes ranging from 1,024 to 16,384. These datasets are sampled with replacements from the DOTmark dataset \cite{schrieber2016dotmark} with an image size of $32\times 32$. From the results in Table~\ref{tab:vectorization}, batch processing achieves significant speedup across all the tested dataset sizes. Note that the speedup from batch processing becomes more pronounced as the dataset size becomes larger. This showcases a nearly 10-fold acceleration from $2.5\times$ for 1,024 images to $23\times$ for 16,384 images.

\begin{table}[t]
    \centering
    
    \caption{(a) Execution time of \alg with/without batch processing on different numbers of images. (b) Mean (SD) of relative error (\%) between \alg and the exact EMD with either $L_1$ or $L_2$ distance on various sizes of images.}
    
    \label{tab:main_table2}
    \renewcommand{\arraystretch}{1}
    \begin{subtable}{1\linewidth}
        \centering
        \scriptsize
        \caption{\label{tab:vectorization}}
        \resizebox{\linewidth}{!}{%
            \begin{tabular}{c|ccccc}
                \toprule
                \multicolumn{1}{c|}{{\textbf{Number}}}& \multicolumn{5}{c}{\textbf{Execution Time (sec)}$\downarrow$} \\
                \multicolumn{1}{c|}{{\textbf{of Images}}}& 1024        & 2048    & 4096  & 8192    & 16384 \\
                \midrule
                w/o Batch. & 52.29   & 137.80    & 278.44   & 558.01  & 1108.32 \\
                w/ Batch.  & 20.92  & 27.25  & 32.64  & 40.05  & 48.19 \\
                Speedup & 2.5$\times$ & 5.1$\times$ & 8.5$\times$ & 14$\times$ & 23$\times$\\
                \bottomrule
            \end{tabular}
        }
    \end{subtable}%
    \renewcommand{\arraystretch}{1}
    
\begin{subtable}{1\linewidth}
        \vspace{7pt}
        \centering
        \scriptsize
        \caption{\label{tab:relative_error}}
        \renewcommand{\arraystretch}{1}
        \resizebox{\linewidth}{!}{%
            \begin{tabular}{c|cccc}
                \toprule
                \multicolumn{1}{c|}{\textbf{\# of Pixels}}& \multicolumn{4}{c}{\textbf{Mean (SD) Relative Error (\%)}$\downarrow$} \\
                \multicolumn{1}{c|}{\textbf{per Image}} & 1024  & 4096      & 9216      & 16384     \\ \midrule
                $L_1$  & 0.22 (0.14)  & 0.67 (0.31) & 0.49 (0.24) & 0.52 (0.37) \\
                $L_2$ & 0.15 (0.07)  & 0.32 (0.11) & 0.37 (0.08) & 0.32 (0.09)\\
                \bottomrule
            \end{tabular}
        }
\end{subtable}

\end{table}

\textbf{Comparison of \texorpdfstring{$L_1$ and $L_2$}{L1 and L2} Distances.} \label{sec:distance} In \alg, the NNS operation needs a distance metric to measure the distances between data points, in order to identify nearest neighbors. This distance metric directly impacts the accuracy and robustness of the approximation results. 

In Table~\ref{tab:relative_error}, we compare the relative error of \alg with respect to the exact EMD to evaluate the $L_1$ and $L_2$ distances on the DOTmark dataset. The relative error is computed as $|R_p - R_p^{opt}| / R_p^{opt}$, where $R_p$ and $R_p^{opt}$ represent the results of \alg and the exact EMD with $L_p$ distance ($p\in \{1,2\}$), respectively. 
For each repeat of this experiment, 10 images are randomly selected for each image size. We summarize the mean and SD of the relative error across 10 repeats to ensure the reliability of our findings.
Table~\ref{tab:relative_error} shows that the $L_2$-based \alg yields a more accurate approximation to the exact EMD than the $L_1$ distance. Further, the $L_2$ distance demonstrates superior stability (smaller SD) over the $L_1$ distance.

\section{Discussion}
{
Though $L_1$ and $L_2$ norms are employed as the ground distances in our experiments, this is not a requirement for \alg. Both our algorithm and theorems are compatible with any valid distance matrix $\mathbf{D}$, thus can support different domain-specific needs. For instance, the Mahalanobis distance takes into account correlations for numerical data, cosine similarity is often preferred for high-dimensional embeddings in text and vision tasks, and geodesic-based distances are often used for manifold-structured data. We plan to investigate how alternative distance metrics affect the performance of NNS-EMD in the future. Another natural direction for future work is to extend \alg to $p$-Wasserstein distances, as well as more general geometric OT problems. Adapting the nearest-neighbor search to handle higher-order distance functions would require new theoretical developments and algorithmic designs, which we plan to explore in the future.}

{Considering that EMD is widely used as a loss function in neural networks training, where differentiability is necessary to enable backpropagation for gradient flow, Sinkhorn-based approximations are often preferred over LP-based exact EMD. Currently, the discrete operations (NNS, sorting, and iterative updates to flow weight) used in our \alg are inherently non-differentiable and pose challenges for backpropagation. As future work, we plan to explore differentiable extensions of \alg for broader applicability.}



\section{Conclusion}
In this work, we introduced \alg, a new computationally efficient EMD approximation method that relies on nearest neighbor search. Empirical evaluations confirm that \alg attains high accuracy with reduced execution time compared to state-of-the-art approximate EMD algorithms. We also presented theoretical analysis on the time complexity and error bounds of \alg. Furthermore, we accelerated \alg through exploiting GPU parallelism. In future work, we plan to investigate the integration of \alg as a loss function within deep learning frameworks, particularly for 3D point cloud applications.



\bibliographystyle{IEEEtran}
\bibliography{main}

\end{document}